\documentclass[a4paper]{article}
\usepackage{RR}
\usepackage{hyperref}
\RRdate{Novembre 2008}

\RRauthor{
Robin Genuer 
  \thanks[sfn]{Universit\'e Paris-Sud, Math\'ematique,
   B\^at.~425, 91405 Orsay, France}%
  \and
Jean-Michel Poggi\thanks{Universit\'e Paris Descartes, France}
\thanksref{sfn}
  \and
Christine Tuleau \thanks{Universit\'e Nice Sophia-Antipolis, France}
}
\authorhead{Genuer \emph{et. al}}
\RRtitle{For\^ets al\'eatoires : remarques m\'ethodologiques}
\RRetitle{Random Forests: some methodological insights}
\titlehead{Random Forests: some methodological insights}
\RRresume{On s'int\'eresse \`a la m\'ethode des for\^ets al\'eatoires 
d'un point de vue m\'ethodologique. Introduite par Leo Breiman en 2001, 
elle est d\'esormais largement utilis\'ee tant en classification qu'en 
r\'egression avec un succ\`es spectaculaire. On vise tout d'abord \`a 
confirmer les r\'esultats exp\'erimentaux, connus mais \'epars, quant 
au choix des param\`etres de la m\'ethode, tant pour les probl\`emes 
dits "standards" que pour ceux dits de "grande dimension" (pour lesquels 
le nombre de variables est tr\`es grand vis \`a vis du nombre d'observations). 
Mais la contribution principale de cet article est d'\'etudier le comportement du score d'importance des variables bas\'e sur les for\^ets al\'eatoires et d'examiner deux probl\`emes classiques de 
s\'election de variables. Le premier est de d\'egager les variables 
importantes \`a des fins d'interpr\'etation tandis que le second, 
plus restrictif, vise \`a se restreindre \`a un sous-ensemble suffisant 
pour la pr\'ediction. La strat\'egie g\'en\'erale proc\`ede en deux 
\'etapes : le classement des variables bas\'e sur les scores d'importance 
suivi d'une proc\'edure d'introduction ascendante s\'equentielle des variables.}

\RRabstract{This paper examines from an experimental perspective random forests,
the increasingly used statistical method for classification and
regression problems introduced by Leo Breiman in 2001. It first aims
at confirming, known but sparse, advice for using random forests and
at proposing some complementary remarks for both standard problems
as well as high dimensional ones for which the number of variables
hugely exceeds the sample size. But the main contribution of this
paper is twofold: to provide some insights about the behavior of the
variable importance index based on random forests and in addition,
to propose to investigate two classical issues of variable
selection. The first one is to find important variables for
interpretation and the second one is more restrictive and try to
design a good prediction model. The strategy involves a ranking of
explanatory variables using the random forests score of importance
and a stepwise ascending  variable introduction strategy.}
\RRmotcle{{\sc For\^ets al\'eatoires, R\'egression, Classification,
Importance des Variables, S\'election des Variables.}}

\RRkeyword{{\sc Random Forests, Regression, Classification,
Variable Importance, Variable Selection.}}
\RRprojets{{\sc Select}}
\RRtheme{\THCog} 
\RCSaclay 

\usepackage{amsmath}
\usepackage{amsfonts}
\usepackage{latexsym}
\usepackage{psfig}
\usepackage{graphicx}

\newtheorem{remark}{Remark}[section]

\begin{document}
\RRNo{6729}
\makeRR   


\section{Introduction}
Random forests (RF henceforth) is a popular and very efficient
algorithm, based on model aggregation ideas, for both classification
and regression problems, introduced by Breiman (2001)
\cite{Breiman01}. It belongs to the family of ensemble methods,
appearing in machine learning at the end of nineties (see for
example Dietterich (1999) \cite{Dietter99} and (2000)
\cite{Dietter00}). Let us briefly recall the statistical framework
by considering a learning set $L=\{(X_1,Y_1),\ldots,(X_n,Y_n)\}$
made of $n$ i.i.d. observations of a random vector $(X,Y)$. Vector
$X=(X^1,...,X^p)$ contains predictors or explanatory variables, say
$X\in\mathbb{R}^p$, and $Y\in\mathcal{Y}$ where $\mathcal{Y}$ is
either a class label or a numerical response. For classification
problems, a classifier $t$ is a mapping $t:\mathbb{R}^p\rightarrow
\mathcal{Y}$ while for regression problems, we suppose that $Y =
s(X) + \varepsilon$ and $s$ is the so-called regression function.
For more background on statistical learning, see Hastie \emph{et
al.} (2001) \cite{Hastie01}. Random forests is a model building
strategy providing estimators of either the Bayes classifier or the
regression function.

The principle of random forests is to combine many binary decision
trees built using several bootstrap samples coming from the learning
sample $L$ and choosing randomly at each node a subset of
explanatory variables $X$. More precisely, with respect to the
well-known CART model building strategy (see Breiman \emph{et al.}
(1984) \cite{Breiman84}) performing a growing step followed by a
pruning one, two differences can be noted. First, at each node, a
given number (denoted by $mtry$) of input variables are randomly
  chosen and the best split is calculated only within this subset.
  Second, no pruning step is performed so all the trees are maximal
  trees.

In addition to CART, another  well-known related tree-based method
must be mentioned: bagging (see Breiman (1996) \cite{Breiman96}).
Indeed random forests with $mtry=p$ reduce simply to unpruned
bagging. The associated R\footnote[1]{see http://www.r-project.org/}
packages are respectively \texttt{randomForest} (intensively used in
the sequel of the paper), \texttt{rpart} and \texttt{ipred} for CART
and bagging respectively (cited here for the sake of completeness).

RF algorithm becomes more and more popular and appears to be very
powerful in a lot of different applications (see for example
D\'{i}az-Uriarte and Alvarez de Andr\'{e}s (2006) \cite{Diaz06} for
gene expression data analysis) even if it is not clearly elucidated
from a mathematical point of view (see the recent paper by Biau
\emph{et al.} (2008) \cite{Biau08} and B\"{u}hlmann, Yu (2002)
\cite{Buhlmann02} for bagging). Nevertheless, Breiman (2001)
\cite{Breiman01} sketches an explanation of the good performance of
random forests related to the good quality of each tree (at least
from the bias point of view) together with the small correlation
among the trees of the forest, where the correlation between trees
is defined as the ordinary correlation of predictions on so-called
 out-of-bag (OOB henceforth) samples. The OOB sample which is the
set of observations which are not used for building the current
tree, is used to estimate the prediction error and
then to evaluate variable importance.\\

\textbf{Tuning method parameters}

It is now classical to distinguish two typical situations depending
on $n$ the number of observations, and $p$ the number of variables:
standard (for $n>>p$) and high dimensional (when $n<<p$). The first
question when someone try to use practically random forests is to
get information about sensible values for the two main parameters of
the method. Essentially, the study carried out in the two papers
\cite{Breiman01} and \cite{Diaz06} give interesting insights but
Breiman focuses on standard problems while D\'{i}az-Uriarte and Alvarez de Andr\'{e}s
concentrate on high dimensional classification ones.

So the first objective of this paper is to give compact information
about selected bench datasets and to examine again the choice of the
method parameters addressing more closely the
different situations. \\

%
%

\textbf{RF variable importance}

The quantification of the variable importance (VI henceforth) is an
important issue in many applied problems complementing variable
selection by interpretation issues. In the linear regression
framework it is examined for example by Gr\"{o}mping (2007)
\cite{Gromp07}, making a distinction between various variance
decomposition based indicators: "dispersion importance", "level
importance" or "theoretical importance" quantifying explained
variance or changes in the response for a given change of each
regressor. Various ways to define and compute using R such
indicators are available (see Gr\"{o}mping (2006) \cite{Gromp06}).

In the random forests framework, the most widely used score of
importance of a given variable is the increasing in mean of the
error of a tree (MSE for regression and misclassification rate for
classification) in the forest when the observed values of this
variable are randomly permuted  in the OOB samples. Often, such
random forests VI is called permutation importance indices in
opposition to total decrease of node impurity measures already
introduced in the seminal book about CART by Breiman \emph{et al.}
(1984) \cite{Breiman84}.

Even if only little investigation is available about RF variable
importance, some interesting facts are collected for classification
problems. This index can be based on the average loss of another
criterion, like the Gini entropy used for growing classification
trees. Let us cite two remarks. The first one is that the RF Gini
importance is not fair in favor of predictor variables with many
categories while the RF permutation importance is a more reliable
indicator (see Strobl \emph{et al.} (2007) \cite{Strobl07}). So we
restrict our attention to this last one. The second one is that it
seems that permutation importance overestimates the variable
importance of highly correlated variables and they propose a
conditional variant (see Strobl \emph{et al.} (2008)
\cite{Strobl08}). Let us mention that, in this paper, we do not
notice such phenomenon. For classification problems, Ben Ishak,
Ghattas (2008) \cite{BenIshak08} and D\'{i}az-Uriarte, Alvarez de
Andr\'{e}s (2006) \cite{Diaz06} for example, use RF variable
importance and note that it is stable for correlated predictors,
scale invariant and stable with respect to small perturbations of
the learning sample. But these preliminary remarks need to be
extended and the recent paper by Archer \emph{et al.} (2008)
\cite{Archer08}, focusing more specifically on the VI topic, do not
answer some crucial questions about the variable importance
behavior: like the importance of a group of variables or its
behavior in presence of
highly correlated variables. This one is the second goal of this paper. \\

\textbf{Variable selection }

Many variable selection procedures are based on the cooperation of
variable importance for ranking and model estimation to evaluate and
compare a family of models. Three types of variable selection
methods are distinguished (see Kohavi \emph{et al.} (1997)
\cite{Kohavi97} and Guyon \emph{et al.} (2003) \cite{Guyon03}):
"filter" for which the score of variable importance does not depend
on a given model design method; "wrapper" which include the
prediction performance in the score calculation; and finally
"embedded" which intricate more closely variable selection and model
estimation.

For non-parametric models, only a small number of methods are
available, especially for the classification case. Let us briefly
mention some of them, which are potentially competing tools. Of
course we must firstly mention the wrapper methods based on VI
coming from CART, see Breiman \emph{et al.} (1984) \cite{Breiman84}
and of course, random forests, see Breiman (2001) \cite{Breiman01}.
Then some examples of embedded methods: Poggi, Tuleau (2006)
\cite{Poggi06} propose a method based on CART scores and using
stepwise ascending procedure with elimination step; Guyon \emph{et
al.} (2002) \cite{Guyon02} (and Rakotomamonjy (2003)
\cite{Rakotomamonjy03}), propose SVM-RFE, a method based on SVM
scores and using descending elimination. More recently, Ben Ishak
\emph{et al.} (2008) \cite{BenIshak08} propose a stepwise variant
while Park \emph{et al.} (2007) \cite{Park07} propose a "LARS" type
strategy (see Efron \emph{et al.} (2004) \cite{Efron04} for
classification problems.

Let us recall that two distinct objectives about variable selection
can be identified: (1) to find important variables highly related to
the response variable for interpretation purpose; (2) to find a
small number of variables sufficient for a good prediction of the
response variable. The key tool for task 1 is thresholding variable
importance while the crucial point for task 2 is to combine variable
ranking and stepwise introduction of variables on a prediction model
building. It could be ascending in order to avoid to select
redundant variables or, for the case $n<<p$, descending first to
reach a classical situation $n\sim p$, and then ascending using the
first strategy, see Fan, Lv (2008) \cite{Fan08}. We propose in this
paper, a two-steps procedure, the first one is common while the
second one
depends on the objective interpretation or prediction. \\

The paper is organized as follows. After this introduction, Section
2 focuses on random forests parameters. Section 3 proposes to study
the behavior of the RF variable importance index. Section 4
investigates the two classical issues of variable selection using
the random forests based score of importance. Section 5 finally
opens discussion about future work.

\section{Selecting method parameters}

\subsection{Experimental framework}

%

\subsubsection{RF procedure}
The R package about random forests is based on the the seminal
contribution of Breiman and Cutler \cite{Breiman05} and is described
in Liaw, Wiener (2002) \cite{Liaw02}. In this paper, we focus on the
\texttt{randomForest} procedure. The two main parameters are $mtry$,
the number of input variables randomly chosen at each split and
$ntree$, the number of trees in the forest\footnote[2]{In all the
paper, $mtry=m$ with $m\in\mathbb{R}$ stands for $mtry=\lfloor
m\rfloor$}.

A third parameter, denoted by $nodesize$, allows to specify the
minimum number of observations in a node. We retain the default
value (1 for classification and 5 for regression) of this parameter
for all of our experimentations, since it is close to the maximal
tree choice.

\subsubsection{OOB error}
In this section, we concentrate on the prediction performance of RF
focusing on out-of-bag (OOB) error (see \cite{Breiman01}). We use
this kind of prediction error estimate for three reasons: the main
is that we are mainly interested in comparing results instead of
assessing models, the second is that it gives fair estimation
compared to the usual alternative test set error even if it is
considered as a little bit optimistic and the last one, but not the
least, is that it is a default output of the procedure. To avoid
unsignificant sampling effects, each OOB errors is actually the mean
of OOB error over $10$ runs.

\subsubsection{Datasets}

We have collected information about the data sets considered in this
paper: the name, the name of the corresponding data structure (when
different), $n$, $p$, the number of classes $c$ in the multiclass
case, a reference, a website or a package.  The two next tables
contain synthetic information while details are postponed in the
Appendix. We distinguish standard and high dimensional situations
and, in addition, the three problems: regression, 2-class
classification and multiclass classification.

Table \ref{tabstand} displays some information about standard
problems datasets: for classification at the top and for regression
at the bottom.

\begin{table}[!ht]
    \begin{center}
    \begin{tabular}{lccc}
Name & Observations & Variables & Classes \\
\hline
Ionosphere & 351 & 34 & 2 \\
Diabetes & 768 & 8 & 2 \\
Sonar & 208 & 60 & 2 \\
Votes & 435 & 16 & 2 \\
Ringnorm & 200 & 20 & 2 \\
Threenorm & 200 & 20 & 2 \\
Twonorm & 200 & 20 & 2 \\
Glass & 214 & 9 & 6 \\
Letters & 20000 & 16 & 26 \\
Sat-images & 6435 & 36 & 6 \\
Vehicle & 846 & 18 & 4 \\
Vowel & 990 & 10 & 11 \\
Waveform & 200 & 21 & 3 \\
\hline
BostonHousing & 506 & 13 & \\
Ozone & 366 & 12 & \\
Servo & 167 & 4 & \\
Friedman1 & 300 & 10 & \\
Friedman2 & 300 & 4 & \\
Friedman3 & 300 & 4 & \\
    \end{tabular}
    \end{center}
    \caption{Standard problems: data sets for classification at the top, and for regression at the bottom}
    \label{tabstand}
\end{table}

Table \ref{tabhigh} displays high dimensional problems datasets: for
classification at the top and for regression at the bottom.

\begin{table}[!ht]
    \begin{center}
    \begin{tabular}{lccc}
Name & Observations & Variables & Classes \\
\hline
Adenocarcinoma & 76 & 9868 & 2 \\
Colon & 62 & 2000 & 2 \\
Leukemia & 38 & 3051 & 2 \\
Prostate & 102 & 6033 & 2 \\
Brain & 42 & 5597 & 5 \\
Breast & 96 & 4869 & 3 \\
Lymphoma & 62 & 4026 & 3 \\
Nci & 61 & 6033 & 8 \\
Srbct & 63 & 2308 & 4 \\
toys data & 100 & 100 to 1000 & 2\\
\hline
PAC & 209 & 467 & \\
Friedman1 & 100 & 100 to 1000 & \\
Friedman2 & 100 & 100 to 1000 & \\
Friedman3 & 100 & 100 to 1000 & \\
    \end{tabular}
    \end{center}
    \caption{High dimensional problems: data sets for classification at the top, and for regression at the bottom}
    \label{tabhigh}
\end{table}

\subsection{Regression}
About regression problems, even if it seems at first inspection that
the seminal paper by Breiman \cite{Breiman01} closes the debate
about good advice, it remains that the experimental results are
about a variant which is not implemented in the universally used R
package. Moreover, except this reference, at our knowledge, no such
a general paper is available, so we develop again the Breiman's
study both for real and simulated data corresponding to the case
$n>>p$ and we provide some additional study on data corresponding to
the case $n<<p$ (such examples typically come from chemometrics).

We observe that the default value of $mtry$ proposed by the R
package is not optimal, and that there is no improvement by using
random forests with respect to unpruned bagging (obtained for
$mtry=p$).

\subsubsection{Standard problems}

\begin{figure}[!ht]
         \begin{center}
         \includegraphics[width=0.7\textwidth]{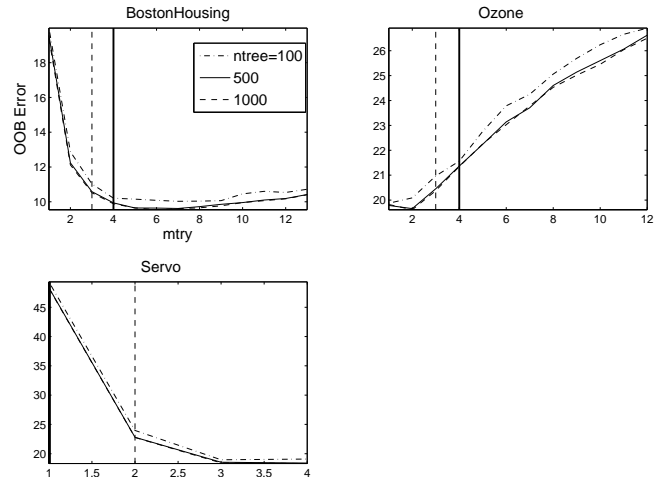}
         \end{center}
         \caption{Standard regression: 3 real data sets}
         \label{regppn}
\end{figure}

\begin{figure}[!ht]
         \begin{center}
         \includegraphics[width=0.7\textwidth]{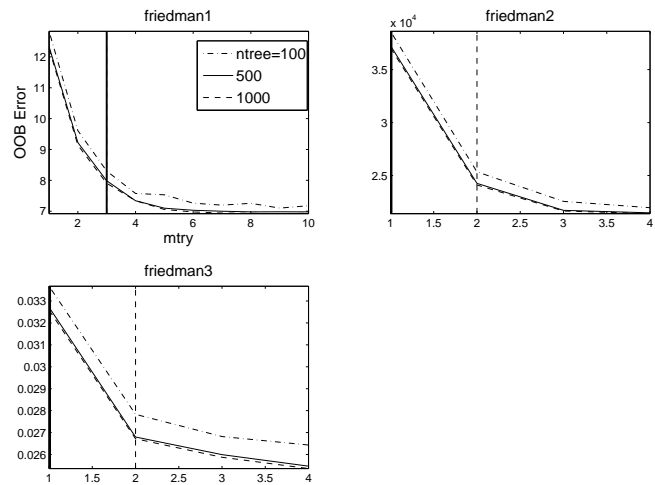}
         \end{center}
         \caption{Standard regression: 3 simulated data sets }
         \label{simulregppn}
\end{figure}

Let us briefly examine standard ($n>>p$) regression datasets. In
Figure \ref{regppn} for real ones and for simulated ones in Figure
\ref{simulregppn}. Each plot gives for $mtry=1$ to $p$ the OOB error
for three different values of $ntree=100, 500$ and $1000$. The
vertical solid line indicates the value $mtry=p/3$, the default
value proposed by the R package for regression problems, the
vertical dashed line being the value $mtry=\sqrt{p}$.

Three remarks can be formulated. First, the OOB error is maximal for
$mtry=1$ and then decreases quickly (except for the ozone dataset,
for   reasons not clearly elucidated), then as soon as
$mtry>\sqrt{p}$, the error remains the same. Second, the choice
$mtry=\sqrt{p}$ gives always lower OOB error than $mtry=p/3$, and
the gain can be important. So the default value proposed by the R
package seems to be often not optimal, especially when $\lfloor
p/3\rfloor=1$. Lastly, the default value $ntree=500$ is convenient,
but a much smaller one $ntree=100$ leads to comparable results.

So, for standard ($n>>p$) regression problems, it seems that there
is no improvement by using random forests with respect to unpruned
bagging (obtained for $mtry=p$).

\subsubsection{High dimensional problems}\label{reghigh}

Let us start with a simulated data set for the high dimensional case
$n<<p$.  This example is built by adding extra noisy variables
(independent and uniformly distributed on $[0,1]$) to the Friedman1
model defined by:
$$Y = 10 \sin(\pi X^1 X^2) + 20 (X^3 - 0.5)^2 + 10 X^4 + 5 X^5 + \epsilon $$
where $X^1,\ldots,X^5$ are independent and uniformly distributed on
$[0,1]$ and $\epsilon\sim \mathcal{N}(0,1)$. So we have $5$
variables related to the response $Y$, the others being noise. We
set $n=100$ and let $p$ vary.

\begin{figure}[!ht]
         \begin{center}
         \includegraphics[width=0.9\textwidth]{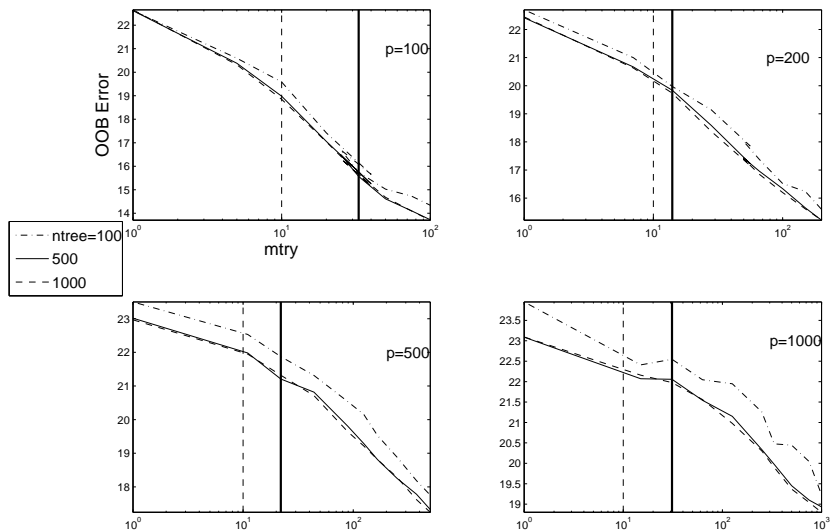}
         \end{center}
         \caption{High dimensional regression simulated data set: Friedman1. The x-axis is in log scale}
         \label{simulregpgn}
\end{figure}

Figure \ref{simulregpgn} contains four plots corresponding to 4
values of $p$ ($100$, $200$, $500$ and $1000$) increasing the
nuisance space dimension. Each plot gives for ten values of $mtry$
($1$, $\sqrt{p}/2$, $\sqrt{p}$, $2\sqrt{p}$, $4\sqrt{p}$, $p/4$,
$p/3$, $p/2$, $3p/4$, $p$) the OOB error for three different values
of $ntree=100, 500$ and $1000$. The x-axis is in log scale and
the vertical solid line indicates $mtry=p/3$ the default value
proposed by the R package for regression, the vertical dashed line
being the value $mtry=\sqrt{p}$.

Let us give four comments. All curves have the same shape: the OOB
error decreases while $mtry$ increases. While $p$ increases, both
OOB errors of unpruned bagging (obtained with $mtry=p$) and random
forests with default value of $mtry$ increase, but unpruned bagging
performs  better than RF (about $25\%$ of improvement). The choice
$mtry=\sqrt{p}$ gives always worse results than those obtained for
$mtry=p/3$. Finally, the default choice $ntree=500$ is convenient,
but a much smaller one $ntree=100$ leads to comparable results.

\begin{figure}[!ht]
         \begin{center}
         \includegraphics[width=0.9\textwidth]{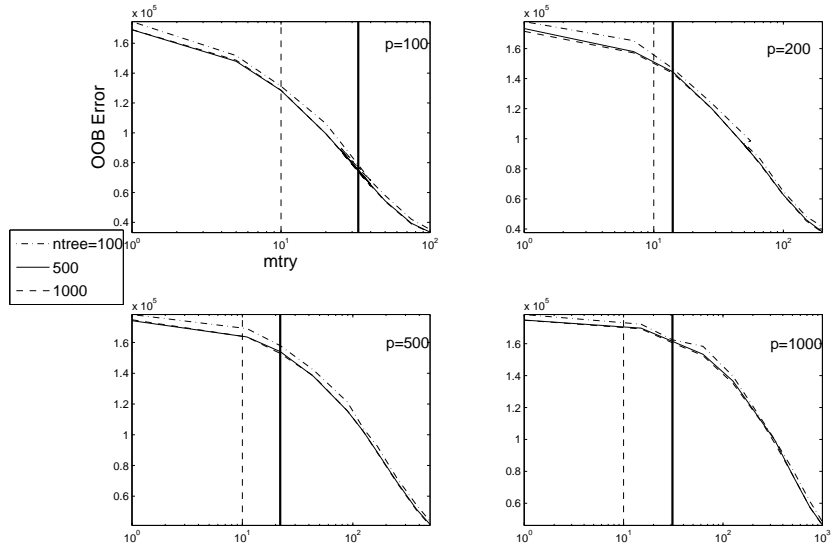}
         \end{center}
         \caption{High dimensional regression simulated data set: Friedman2. The x-axis is in log scale}
         \label{friedman2}
\end{figure}
\begin{figure}[!ht]
         \begin{center}
         \includegraphics[width=0.9\textwidth]{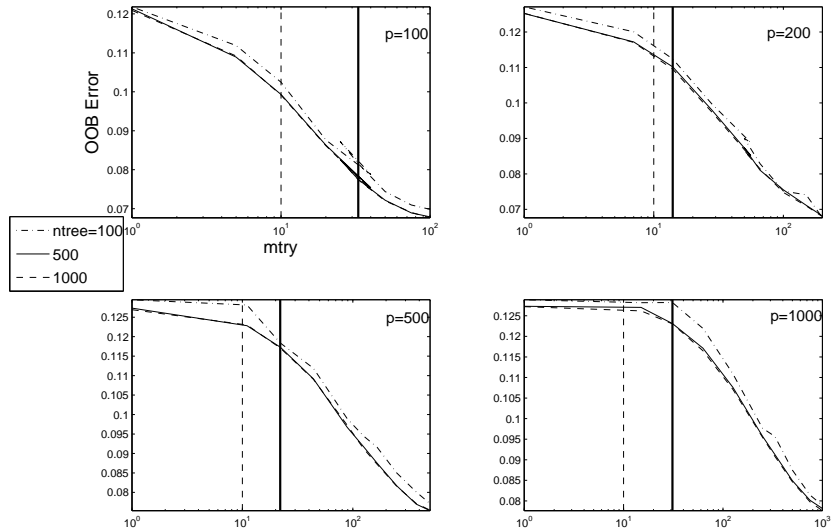}
         \end{center}
         \caption{High dimensional regression simulated data set: Friedman3. The x-axis is in log scale}
         \label{friedman3}
\end{figure}

Figure \ref{friedman2} and \ref{friedman3} show the results of the
same study for the Friedman2 and Friedman3 models. The previous
comments remain valid. Let us just note that the difference between
unpruned bagging and random forests with $mtry$ default value is
even more pronounced for these two problems.

\begin{figure}[!ht]
         \begin{center}
         \includegraphics[width=0.5\textwidth]{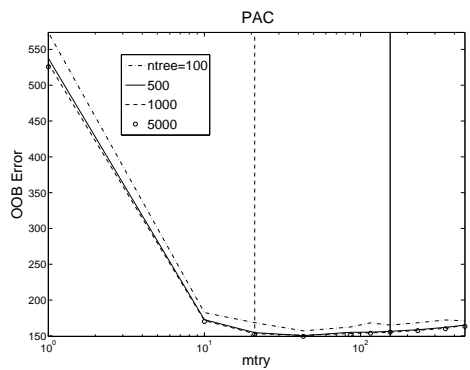}
         \end{center}
         \caption{High dimensional regression: PAC data. The x-axis is in log scale}
         \label{regpgn}
\end{figure}

To end, let us now examine the high dimensional real data set PAC.
Figure \ref{regpgn}  gives for same ten values of $mtry$ the OOB
error for four different values of $ntree=100, 500, 1000$ and $5000$
(x-axis is in log scale). The general behavior is similar except for
the shape: as soon as $mtry>\sqrt{p}$, the error remains the same
instead of still decreasing. The difference of the shape of the
curves between simulated and real datasets can be explained by the
fact that, in simulated datasets we considered, the number of true
variables is very small compared to the total number of variables.
One may expect that in real datasets, the proportion of true
variables is larger.

So, for high dimensional ($n<<p$) regression problems, unpruned
bagging seems to perform  better than random forests and the
difference can be large.


\subsection{Classification}
About standard classification problems, we check that Breiman's
conclusions remain valid for the considered variant and that the
$mtry$ default value proposed in the R package is good. However for
high dimensional classification problems, we observe that larger
values of $mtry$ give sometimes much better results.

\subsubsection{Standard problems}

For classification problems for which $n>>p$, again the paper by
Breiman is interesting and we just quickly check the conclusions.

\begin{figure}[!ht]
         \begin{center}
         \includegraphics[width=0.9\textwidth]{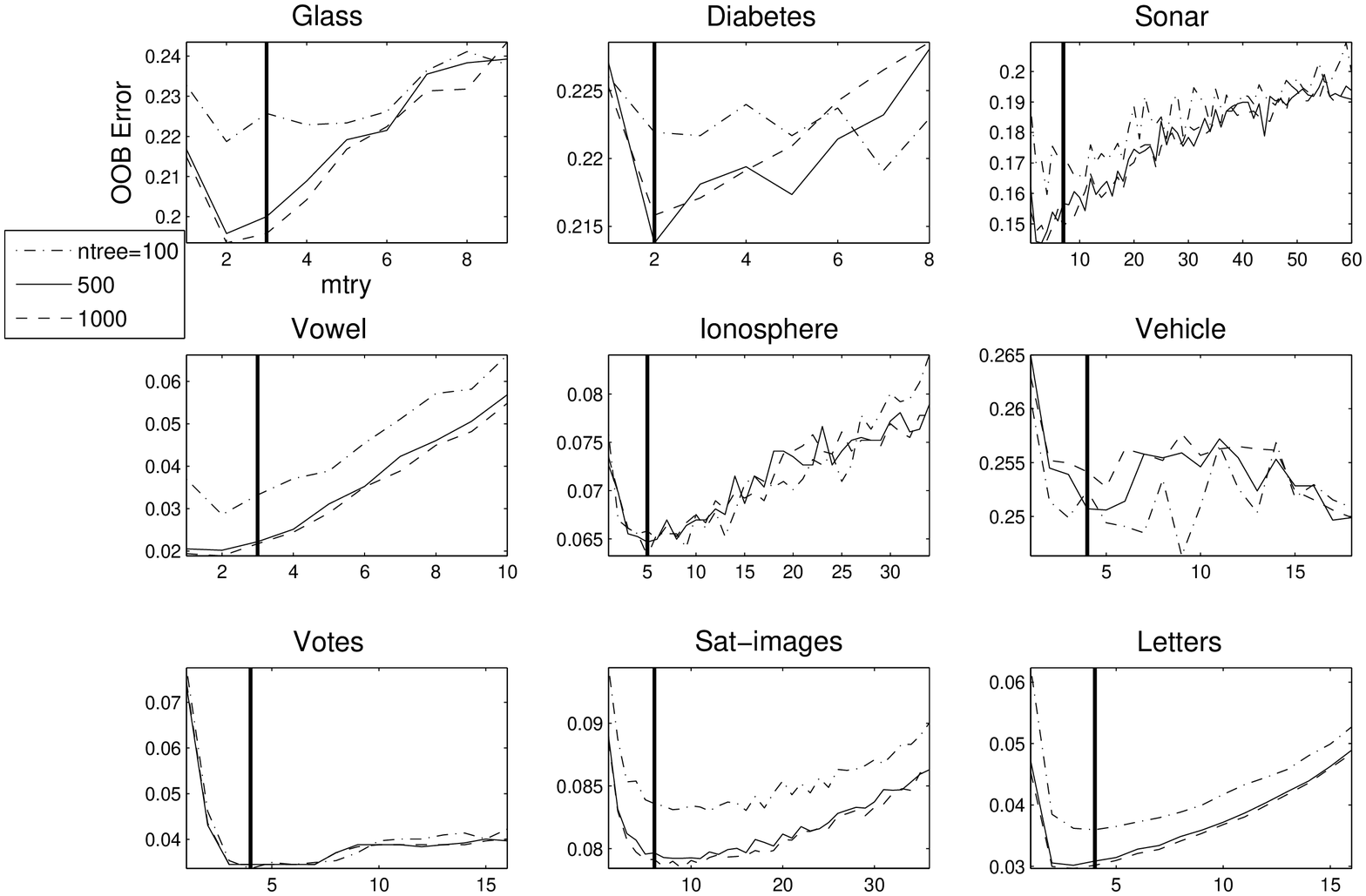}
         \end{center}
         \caption{Standard classification: 9 real data sets}
         \label{ppn}
\end{figure}

Let us first examine in Figure \ref{ppn} standard ($n>>p$)
classification real data sets. Each plot gives for $mtry=1$ to $p$
the OOB error for three different values of $ntree=100, 500$ and
$1000$. The vertical solid line indicates the value $mtry=\sqrt{p}$,
the default value proposed by the R package for classification.

Three remarks can be formulated. The default value $mtry=\sqrt{p}$
is convenient for all the
  examples. The default value $ntree=500$ is sufficient and a much smaller
  one $ntree=100$ is not convenient and can leads to significantly larger
  errors. The general shape is the following: the
  errors for $mtry=1$ and for $mtry=p$ (corresponding to the
  unpruned bagging) are of the same "large" order of magnitude and the
  minimum is reached for the value $\sqrt{p}$. The gain can be
  about 30 or 50\%.

So, for these 9 examples, the default value proposed by the R package is quite optimal.

\begin{figure}[!ht]
         \begin{center}
         \includegraphics[width=0.7\textwidth]{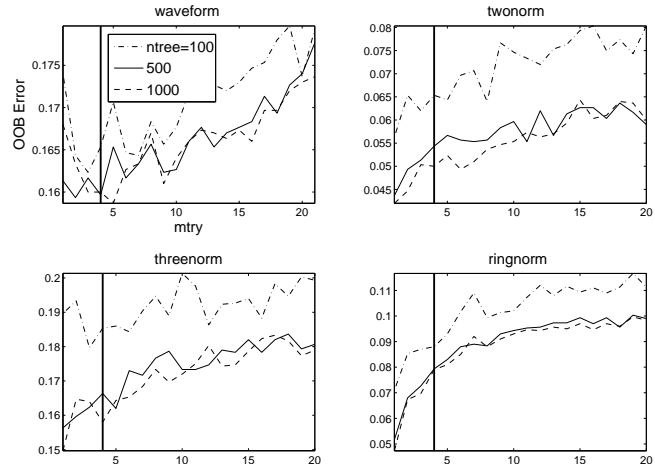}
         \end{center}
         \caption{Standard classification: 4 simulated data sets}
         \label{simulppn}
\end{figure}

Let us now examine in Figure \ref{simulppn} standard ($n>>p$)
classification simulated datasets. As it can be seen, $ntree=500$ is
sufficient and, except for the ringnorm already pointed out as a
somewhat special dataset (see Cutler, Zhao (2001) \cite{Cut01}) the
value $mtry=\sqrt{p}$ is good. Here, the general shape of the error
curve is quite different compared to real datasets: the error
increases with $mtry$. So for these four examples, the smaller
$mtry$, the better.

\subsubsection{High dimensional problems}\label{classhigh}

\begin{figure}[!ht]
         \begin{center}
         \includegraphics[width=0.9\textwidth]{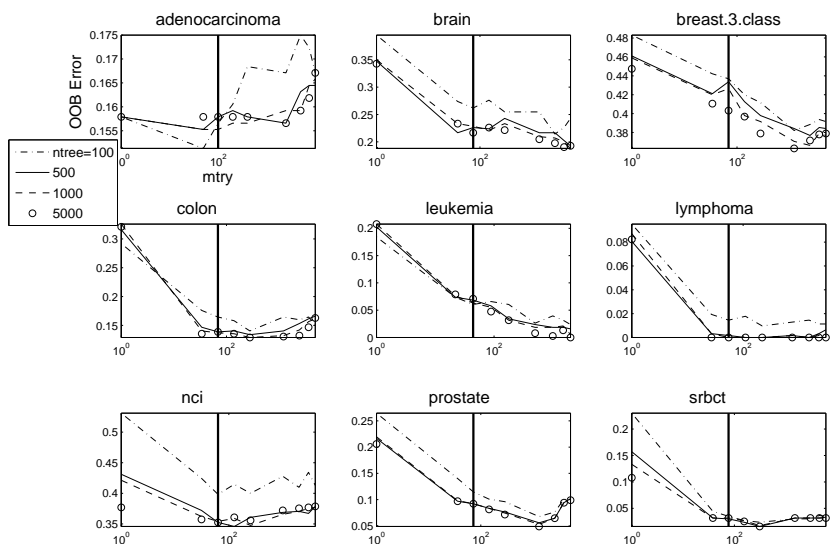}
         \end{center}
         \caption{High dimensional classification: 9 real data sets. The x-axis is in log scale}
         \label{pgn}
\end{figure}

Let us now consider the case $n<<p$ for which D\'{i}az-Uriarte and
Alvarez de Andr\'{e}s (2006) \cite{Diaz06} give numerous advice. We
complete the study by trying larger values of $mtry$, which give
interesting results. One can found in Figure \ref{pgn} the OOB
errors for nine high dimensional real datasets. Each plot gives for
nine values of $mtry$ ($1$, $\sqrt{p}/2$, $\sqrt{p}$, $2\sqrt{p}$,
$4\sqrt{p}$, $p/4$, $p/2$, $3p/4$, $p$) the OOB error for four
different values of $ntree=100, 500, 1000$ and $5000$. The x-axis is
in log scale. The vertical solid line indicates the default value
proposed by the R package $mtry=\sqrt{p}$.

Again the default value $ntree=500$ is sufficient, and at the
contrary the value
  $ntree=100$ can leads to significantly larger errors. The general
  shape is the following: it decreases in general and the
minimum   value is obtained or is close to the one reached using
$mtry=p$ (corresponding to the
  unpruned bagging). The difference with standard problems is notable,
  the reason is that when $p$ is large, $mtry$ must be sufficiently
  large in order to have a high probability to capture important variables (that is
variables highly related to the response) for defining
  the splits of the RF. In addition, let us mention that the default value $mtry=\sqrt{p}$ is still reasonable
  from the OOB error viewpoint but of course, since $\sqrt{p}$ is small with respect to $p$, it
  is a very attractive value from a computational perspective (notice that the trees are
  not too deep since $n$ is not too large).

Let us examine a simulated dataset for the case $n<<p$, introduced
by Weston \emph{et al.} (2003) \cite{Weston03}, called ``toys data''
in the sequel. It is an equiprobable two-class problem,
$Y\in\{-1,1\}$, with $6$ true variables, the others being some
noise. This example is interesting since it constructs two near
independent groups of 3 significant variables (highly, moderately
and weakly correlated with response $Y$) and an additional group of
noise variables, uncorrelated with $Y$. A forward reference to the
plots on the left side of Figure \ref{ivb1} allow to see the
variable importance picture and to note that the importance of the
variables 1 to 3 is much higher than the one of variables 4 to 6.
More precisely, the model is defined through the conditional
distribution of the $X^i$ for $Y=y$:
\begin{itemize}
 \item for $70\%$ of data, $X^i\sim y\mathcal{N}(i,1)$ for $i=1,2,3$ and
$X^i\sim y\mathcal{N}(0,1)$ for $i=4,5,6$.
 \item for the $30\%$ left, $X^i\sim y\mathcal{N}(0,1)$ for $i=1,2,3$ and
$X^i\sim y\mathcal{N}(i-3,1)$ for $i=4,5,6$.
 \item the other variables are noise, $X^i\sim \mathcal{N}(0,1)$
for $i=7,\ldots,p$.
\end{itemize}
After simulation, obtained variables are standardized. Let us fix
$n=100$.

\begin{figure}[!ht]
         \begin{center}
         \includegraphics[width=0.7\textwidth]{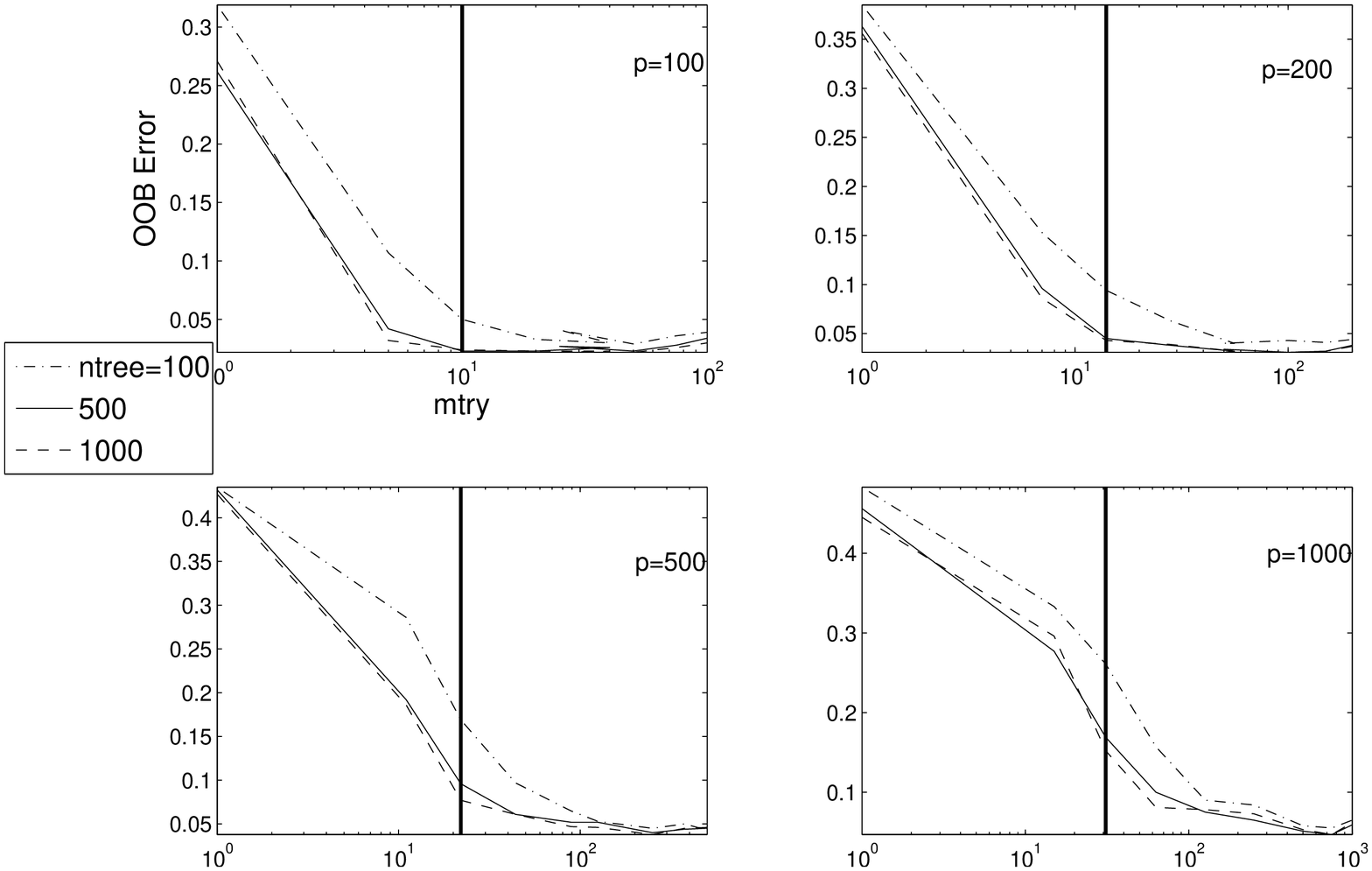}
         \end{center}
         \caption{High dimensional classification simulated data set: toys data for 4 values of $p$. The x-axis is in log scale}
         \label{simulpgn}
\end{figure}

The plots of Figure \ref{simulpgn} are organized as previously, four
values of $p$ are considered: $100$, $200$, $500$ and $1000$
corresponding to increasing  nuisance space dimension. For $p=100$
and $p=200$, the error decreases hugely until $mtry$ reaches
$\sqrt{p}$ and then remains constant, so the default values work
well and perform as well as unpruned bagging, even if the true
dimension $\tilde{p}=6<<p$. For larger values of $p$ ($p\geq 500$),
the shape of the curve is close to the one for high dimensional real
data sets (the error decreases and the minimum is reached when
$mtry=p$). Whence, the error reached by using random forests with
default $mtry$ is about $70\%$ to $150\%$ larger than the error
reached by unpruned bagging which is close to $3\%$ for all the
considered values of $p$.

Finally, for high dimensional classification problems, our
conclusion is that it may be worthwhile to choose $mtry$ larger
than the default value $\sqrt{p}$.

After this section focusing on the prediction performance, let us now
focus on the second attractive feature of RF: the variable
importance index.

\section{Variable importance}

The quantification of the variable importance (abbreviated VI) is a
crucial issue not only for ranking the variables before a stepwise
estimation model but also to interpret data and understand
underlying phenomenons in many applied problems.

In this section, we examine the RF variable importance behavior
according to three different issues. The first one deals with the
sensitivity to the sample size $n$ and the number of variables $p$.
The second examines the sensitivity to method parameters $mtry$ and
$ntree$. The last one deals with the variable importance of a group
of variables, highly correlated or poorly correlated together with
the problem of correct identification of irrelevant variables.

As a result, a good choice of parameters of RF can help to better discriminate
between important and useless variables. In addition, it can
increase the stability of VI scores.

To illustrate this discussion, let us consider the toys data
introduced in Section \ref{classhigh} and compute the variable
importance. Recall that only the first $6$ variables are of interest
and the others are noise.

\begin{remark}
Let us mention that variable importance is computed conditionally to
a given realization even for simulated datasets. This choice which
is criticizable if the objective is to reach a good estimation of an
underlying constant, is consistent with the idea of staying as close
as possible to the experimental situation dealing with a given
dataset. In addition, the number of permutations of the observed
values in the OOB sample, used to compute the score of importance is
set to the default value $1$.
\end{remark}

\subsection{Sensitivity to $n$ and $p$}

\begin{figure}[!ht]
         \begin{center}
         \includegraphics[width=0.7\textwidth]{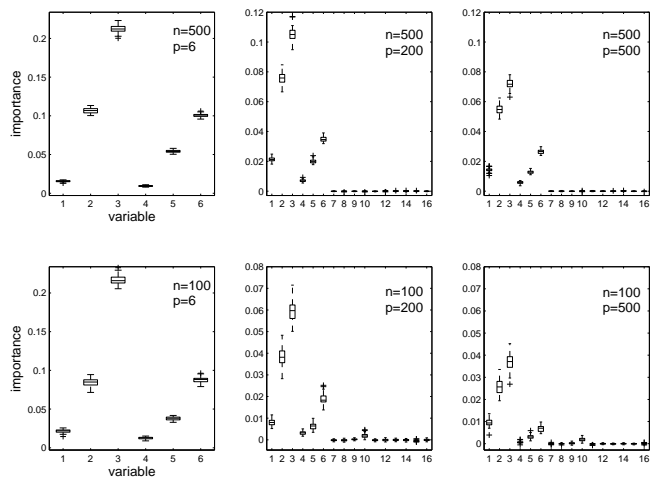}
         \end{center}
         \caption{Variable importance sensitivity to $n$ and $p$ (toys data)}
         \label{ivb1}
\end{figure}

Figure \ref{ivb1} illustrates the behavior of variable importance
for several values of $n$ and $p$. Parameters $ntree$ and $mtry$ are
set to their default values. Boxplots are based on $50$ runs of the
RF algorithm and for visibility, we plot the variable importance
only for a few variables.

On each row, the first plot is the reference one for which we
observe a convenient picture of the relative importance of the
initial variables. Then, when $p$ increases tremendously, we try to
check if: (1) the situation between the two groups remains readable;
  (2) the situation within each group is stable;
  (3) the importance of the additional dummy variables is close to 0.

  The situation $n=500$ (graphs at the top of the figure)
corresponds to an ``easy'' case, where a lot of data are available
and $n=100$ (graphs at the bottom) to a harder one. For each value
of $n$, three values of $p$ are considered: $6,200$ and $500$.  When
$p=6$ only the $6$ true variables are present. Then two very
difficult situations are considered: $p=200$ with a lot of noisy
variables and $p=500$ is even harder. Graphs are truncated after the
$16$th variable for readability (importance of noisy variables left
are the same order of magnitude as the last plotted).

Let us comment on graphs on the first row ($n=500$). When $p=6$ we
obtain concentrated boxplots and the order is clear, variables $2$
and $6$ having nearly the same importance. When $p$ increases, the
order of magnitude of importance decreases. The order within the two
groups of variables ($1,2,3$ and $4,5,6$) remains the same, while
the overall order is modified (variable $6$ is now less important
than variable $2$). In addition, variable importance is more
unstable for huge values of $p$. But what is remarkable is that all
noisy variables have a zero VI. So one can easily recover variables
of interest.

In the second row($n=100$), we note a greater instability since the
number of observations is only moderate, but the variable ranking
remains quite the same. What differs is that in the difficult
situations ($p=200,500$) importance of some noisy variables
increases, and for example variable $4$ cannot be highlighted from
noise (even variable $5$ in the bottom right graph). This is due to
the decreasing behavior of VI with $p$ growing, coming from the fact
that when $p=500$ the algorithm randomly choose only $22$ variables
at each split (with the $mtry$ default value). The probability of
choosing one of the $6$ true variables is really small and the less
a variable is chosen, the less it can be considered as important.

In addition, let us remark that the variability of VI is large for
true variables with respect to useless ones. This remark can be used
to build some kind of test for VI (see Strobl \emph{et al.} (2007)
\cite{Strobl07}) but of course ranking is better suited for variable
selection.

We now study how this VI index behaves when changing values of the
main method parameters.

\subsection{Sensitivity to $mtry$ and $ntree$}

\begin{figure}[!ht]
         \begin{center}
         \includegraphics[width=0.7\textwidth]{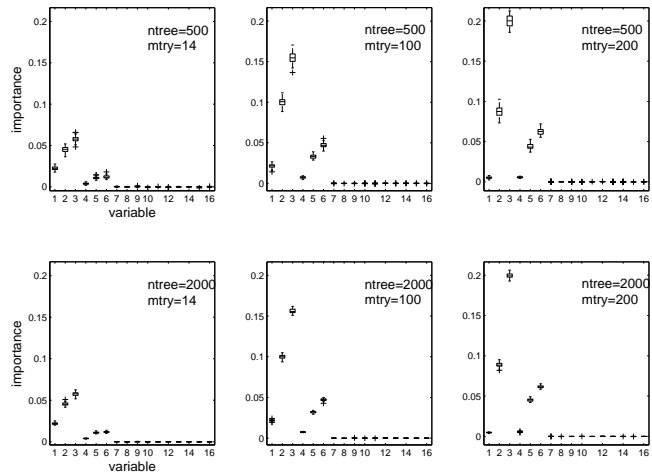}
         \end{center}
         \caption{Variable importance sensitivity to $mtry$ and $ntree$ (toys data)}
         \label{ivb2}
\end{figure}

The choice of $mtry$ and $ntree$ can be important for the VI
computation. Let us fix $n=100$ and $p=200$. In Figure \ref{ivb2} we
plot variable importance obtained using three values of $mtry$ ($14$
the default, $100$ and $200$) and two values of $ntree$ ($500$ the
default, and $2000$).

The effect of taking a larger value for $mtry$ is obvious. Indeed
the magnitude of VI is more than doubled starting from $mtry=14$ to
$mtry=100$, and it again increases whith $mtry=200$. The effect of
$ntree$ is less visible, but taking $ntree=2000$ leads to better
stability. What is interesting in the bottom right graph is that we
get the same order for all true variables in every run of the
procedure. In top left situation the mean OOB error rate is about
$5\%$ and in the bottom right one it is $3\%$. The gain in error may
not be considered as large, but what we get in VI is interesting.

\subsection{Sensitivity to highly correlated predictors}

Let us address an important issue: how does variable importance
behave in presence of several highly correlated variables? We take
as basic framework the previous context with $n=100$, $p=200$,
$ntree=2000$ and $mtry=100$. Then we add to the dataset highly
correlated replications of some of the 6 true variables. The
replicates are inserted between the true variables and the useless
ones.

\begin{figure}[!ht]
         \begin{center}
         \includegraphics[width=0.7\textwidth]{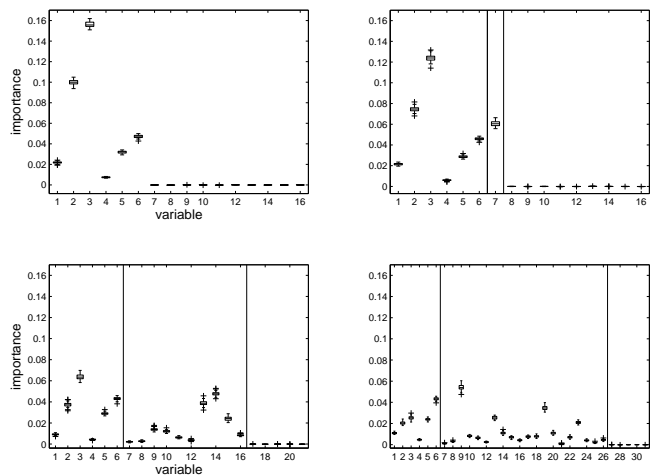}
         \end{center}
         \caption{Variable importance of a group of correlated variables (augmented toys data)}
         \label{ivb3}
\end{figure}

The first graph of Figure \ref{ivb3} is the reference one: the
situation is the same as previously. Then for the three other cases,
we simulate $1$, $10$ and $20$ variables with a correlation of $0.9$
with variable $3$ (the most important one). These replications are
plotted between the two vertical lines.

The magnitude of importance of the group $1,2,3$ is steadily
decreasing when adding more replications of variable $3$. On the
other hand, the importance of the group $4,5,6$ is unchanged. Notice
that the importance is not divided by the number of replications.
Indeed in our example, even with $20$ replications the maximum
importance of the group containing variable $3$ (that is variable
$1,2,3$ and all replications of variable $3$) is only three times
lower than the initial importance of variable $3$. Finally, note
that even if some variables in this group have low importance, they
cannot be confused with noise.

\begin{figure}[!ht]
         \begin{center}
         \includegraphics[width=0.7\textwidth]{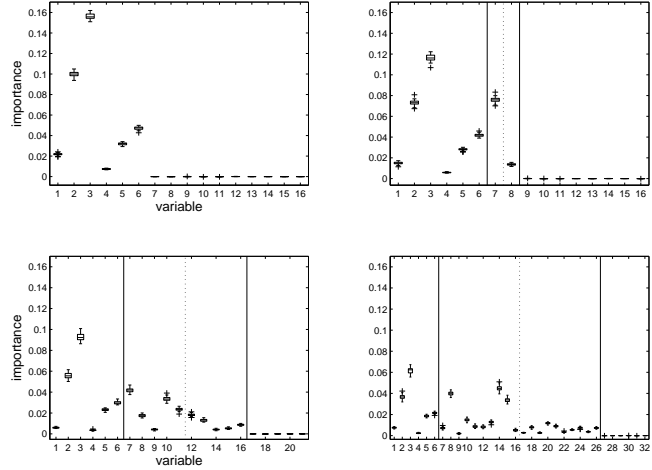}
         \end{center}
         \caption{Variable importance of two groups of correlated variables (augmented toys data)}
         \label{ivb4}
\end{figure}

Let us briefly comment on similar experiments (see Figure
\ref{ivb4}) but perturbing the basic situation not only by
introducing highly correlated versions of the third variable but
also of the sixth, leading to replicate the most important of each
group.

Again, the first graph is the reference one. Then we simulate $1$,
$5$ and $10$ variables of correlation about $0.9$ with variable $3$
and the same with variable $6$. Replications of variable $3$ are
plotted between the first vertical line and the dashed line, and
replications of variable $6$ between the dashed line and the second
vertical line.

The magnitude of importance of each group ($1,2,3$ and $4,5,6$
respectively) is steadily decreasing when adding more replications.
The relative importance between the two groups is preserved. And the
relative importance between the two groups of replications is of the
same order than the one between the two initial groups.

\subsection{Prostate data variable importance}

\begin{figure}[!ht]
         \begin{center}
         \includegraphics[width=0.9\textwidth]{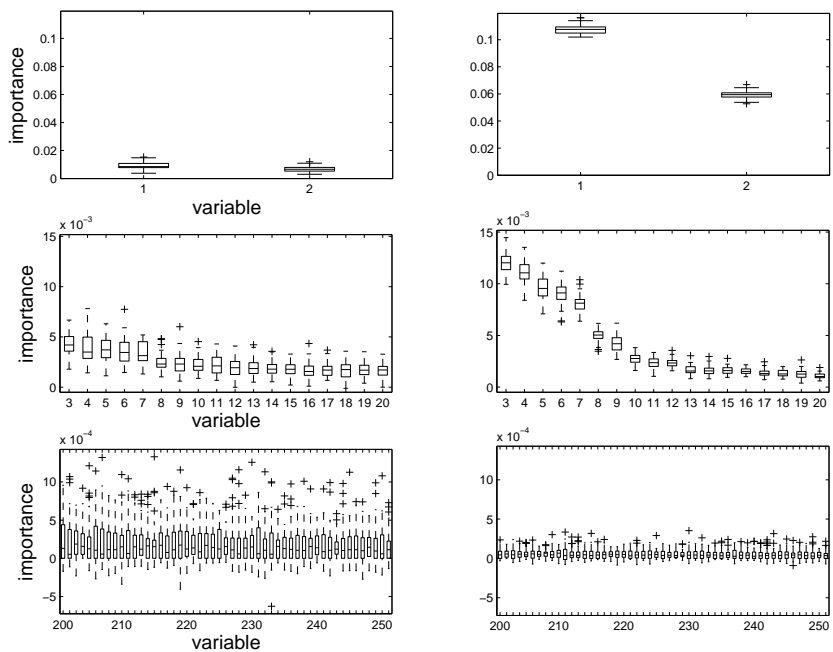}
         \end{center}
         \caption{Variable importance for Prostate data (using $ntree=2000$ and $mtry=p/3$, on the right and using default values
on the left)}
         \label{ivbr}
\end{figure}

To end this section, we illustrate the behavior of variable
importance on a high dimensional real dataset: the microarray data
called Prostate. The global picture is the following: two variables
hugely important, about twenty moderately important variables and
the others of small importance. So, more precisely Figure \ref{ivbr}
compares VI obtained for parameters set to their default values
(graphs of the left column) and those obtained for $ntree=2000$ and
$mtry=p/3$ (graphs of the right column).

Let us comment on Figure \ref{ivbr}. For the two most important
variables (first row), the magnitude
 of importance obtained with $ntree=2000$ and $mtry=p/3$
is much larger than to the one obtained with default values. In the
second row, the increase of magnitude is still noticeable from the
third to the 9th most important variables and from the 10th to the
20th most important variables, VI is quite the same for the two
parameter choices. In the third row, we get VI closer to zero for
the variables with $ntree=2000$ and $mtry=p/3$ than with default
values. In addition, note that for the less important variables,
boxplots are larger for default values, especially
 for unimportant variables (from the 200th to the 250th).

\section{Variable selection}

\subsection{Procedure}

\subsubsection{Principle}

We distinguish two variable selection objectives:
\begin{enumerate}
  \item to find important variables highly related to the response variable for interpretation purpose;
  \item to find a small number of variables sufficient to a good prediction of the response
  variable.
\end{enumerate}
The first is to magnify all the important variables, even with high
redundancy, for interpretation purpose and the second is to find a
sufficient parsimonious set of important variables for prediction.

Two earlier works must be cited: D\'{i}az-Uriarte, Alvarez de
Andr\'{e}s (2006) \cite{Diaz06} and Ben Ishak, Ghattas (2008)
\cite{BenIshak08}.

D\'{i}az-Uriarte, Alvarez de Andr\'{e}s propose a strategy based on
recursive elimination of variables. More precisely, they first
compute RF variable importance. Then, at each step, they eliminate
the $20\%$ of the variables having the smallest importance and build
a new forest with the remaining variables. They finally select the
set of variables leading to the smallest OOB error rate. The
proportion of variables to eliminate is an arbitrary parameter of
their method and does not depend on the data.

Ben Ishak, Ghattas choose an ascendant strategy based on a
sequential introduction of variables. First, they compute some
SVM-based variable importance. Then, they build a sequence of SVM
models invoking at the beginning the $k$ most important variables,
by step of 1. When $k$ becomes too large, the additional variables
are invoked by packets. They finally select the set of variables
leading to the model of smallest error rate. The way to introduce
variables is not data-driven since it is fixed before running the
procedure. They also compare their procedure  with a similar one
using RF instead of SVM.

We propose the following two-steps procedure, the first one is
common while the second one depends on the objective:
\begin{enumerate}
  \item Preliminary elimination and ranking:
  \begin{itemize}
    \item Compute the RF scores of importance, cancel the variables
    of small importance;
    \item Order the $m$ remaining variables in decreasing order of
    importance.
  \end{itemize}
  \item Variable selection:
  \begin{itemize}
    \item For \emph{interpretation}: construct the nested collection of
    RF models involving the $k$ first variables, for $k=1$ to
    $m$ and select the variables involved in the model leading to the smallest OOB
    error;
    \item For \emph{prediction}: starting from the ordered variables retained for interpretation,
    construct an ascending sequence of RF models, by invoking
and testing the variables stepwise. The variables of the last model
are selected.
  \end{itemize}
\end{enumerate}

Of course, this is a sketch of procedure and more details are needed
to be effective. The next paragraph answer this point but we
emphasize that we propose an heuristic strategy which is not
supported by specific model hypotheses but based on data-driven
thresholds to take decisions.

\begin{remark}
Since we want to treat in an unified way all the situations, we will
use for finding prediction variables the somewhat crude strategy
previously defined. Nevertheless, starting from the set of variables
selected for interpretation (say of size $K$), a better strategy
could be to examine all, or at least a large part, of the $2^K$
possible models and to select the variables of the model minimizing
the OOB error. But this strategy becomes quickly unrealistic for
high dimensional problems so we prefer to experiment a strategy
designed for small $n$ and large $K$ which is not conservative and
even possibly leads to select fewer variables.
\end{remark}

\subsubsection{Starting example}

\begin{figure}[!ht]
         \begin{center}
         \includegraphics[width=0.7\textwidth]{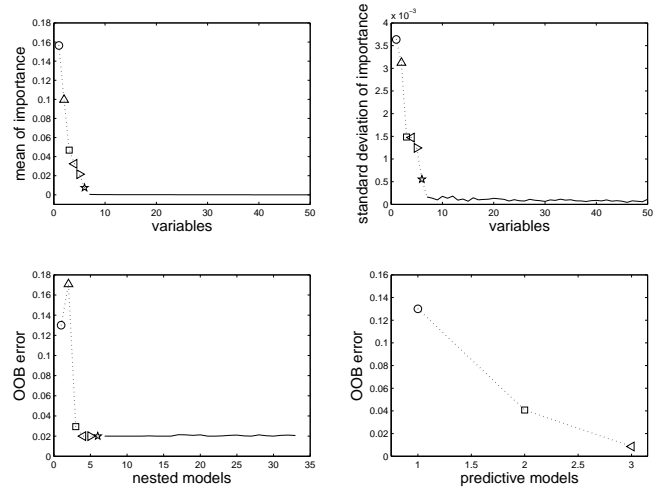}
         \end{center}
         \caption{Variable selection procedures for interpretation and prediction for toys data}
         \label{varselect}
\end{figure}

To both illustrate and give more details about this procedure, we
apply it on a simulated learning set of size $n=100$ from the
classification toys data model (see Section \ref{classhigh}) with
$p=200$. The results are summarized in Figure \ref{varselect}. The
true variables ($1$ to $6$) are respectively represented by
($\rhd,\bigtriangleup,\circ,\star,\lhd,\square$). We compute, thanks
to the learning set, $50$ forests with $ntree=2000$ and $mtry=100$,
which are values of the main parameters previously considered as
well adapted for VI calculations.

Let us detail the main stages of the procedure together with the
results obtained on toys data:
\begin{itemize}
 \item First we rank the variables by sorting the VI in descending order.

The result is drawn on the top left graph for the $50$ most important
variables (the other noisy variables having an importance very close
to zero too). Note that true variables are significantly more
important than the noisy ones.

 \item We keep this order in mind and plot the corresponding standard
deviations of VI. We use this graph to estimate some threshold for
importance, and we keep only the variables of importance exceeding
this level. More precisely, we select the threshold as the minimum
prediction value given by a CART model fitting this curve. This rule
is, in general conservative and leads to retain more variables than
necessary in order to make a careful choice later.

The standard deviations of VI can be found in the top right graph.
We can see that true variables standard deviation
 is large compared to the noisy variables
one, which is close to zero. The threshold leads to retain $33$
variables.

 \item Then, we compute OOB error rates of random forests (using default
 parameters) of the nested models starting from the one
 with only the most important variable, and ending with the one involving all important
variables kept previously. The variables of the model leading to the
smallest OOB error are selected.

Note that in the bottom left graph the error decreases quickly and
reaches its minimum when the first $4$ true variables are included
in the model. Then it remains constant. We select the model
containing $4$ of the $6$ true variables. More precisely, we select
the variables involved in the model \emph{almost} leading to the
smallest OOB error, \emph{i.e.} the first model \emph{almost}
leading to the minimum. The actual minimum is reached with 24
variables.

The expected behavior is non-decreasing as soon as all the "true"
variables have been selected. It is then difficult to treat in a
unified way nearly constant of or slightly increasing. In fact, we
propose to use an heuristic rule similar to the 1 SE rule of Breiman
\emph{et al.} (1984) \cite{Breiman84} used for selection in the
cost-complexity pruning procedure.

 \item We perform a sequential variable introduction with testing: a variable
 is added only if the error gain exceeds a threshold. The idea is that the error decrease
 must be significantly greater than the average variation obtained by adding
 noisy variables.

The bottom right graph shows the result of this step, the final model for
prediction purpose involves only variables $3$, $6$ and $5$. The
threshold is set to the mean of the absolute values
 of the first order differentiated errors between the model with $5$ variables
 (the first model
after the one we selected for interpretation, see the bottom left graph) and
the last one.

\end{itemize}

It should be noted that if one wants to estimate the prediction
error, since ranking and selection are made on the same set of
observations, of course an error evaluation on a test set or using a
cross validation scheme should be preferred. It is taken into
account in the next section when our results are compared to others.

To evaluate fairly the different prediction errors, we prefer here
to simulate a test set of the same size than the learning set. The
test error rate with all (200) variables is about $6\%$ while the
one with the $4$ variables selected for interpretation is about
$4.5\%$, a little bit smaller. The model with prediction variables
$3$, $6$ and $5$ reaches an error of $1\%$. Repeating the global
procedure $10$ times on the same data always gave the same
interpretation set of variables and the same prediction set, in the
same order.

\subsubsection{Highly correlated variables}

Let us now apply the procedure on toys data with replicated
variables: a first group of variables highly correlated with
variable $3$ and a second one replicated from variable $6$ (the most
important variable of each group). The situations of interest are
the same as those considered to produce Figure \ref{ivb4}.

\begin{table}[!ht]
         \begin{center}
     \begin{tabular}{c|c|c}
number of replications & interpretation set & prediction set \\
\hline
1  & 3 $7^3$ 2 6 5                                    & 3 6 5\\
5  & 3  2  $7^3$ $10^3$  6 $11^3$  5 $12^6$               & 3 6 5 \\
10 & 3 $14^3$  $8^3$  2 $15^3$  6  5 $10^3$ $13^3$ $20^6$ & 3  6  5 $10^3$
     \end{tabular}
         \end{center}
         \caption{Variable selection procedure in presence of highly correlated variables (augmented toys data)}
         \label{tabvarselect}
\end{table}

Let us comment on Table \ref{tabvarselect}, where the expression
$i^j$ means that variable $i$ is a replication of variable $j$.

Interpretation sets do not contain all variables of interest.
Particularly we hardly keep replications of variable $6$. The reason
is that even before adding noisy variables to the model the error
rate of nested models do increase (or remain constant): when several highly correlated
variables are added, the bias remains the same while the variance
increases. However the prediction sets are satisfactory: we always
highlight variables $3$ and $6$ and at most one correlated variable
with each of them.

 Even if all the variables of interest do
not appear in the interpretation set, they always appear in the
first positions of our ranking according to importance. More
precisely the $16$ most important variables in the case of $5$
replications are: ($3$   $2$ $7^3$ $10^3$ $6$ $11^3$ $5$  $12^6$
$8^3$  $13^6$
  $16^6$   $1$  $15^6$  $14^6$   $9^3$   $4$), and the $26$
most important variables in the case of $10$ replications are: ($3$
$14^3$  $8^3$  $2$ $15^3$  $6$  $5$ $10^3$ $13^3$ $20^6$ $21^6$
$11^3$ $12^3$ $18^6$ $1$ $24^6$ $7^3$ $26^6$ $23^6$ $16^3$ $25^6$
$22^6$ $17^6$ $19^6$ $4$ $9^3$).
 Note that the order of the true variables ($3$ $2$ $6$ $5$ $1$ $4$)
remains the same in all situations.

\subsection{Classification}

\subsubsection{Prostate data}

\begin{figure}[!ht]
         \begin{center}
         \includegraphics[width=0.7\textwidth]{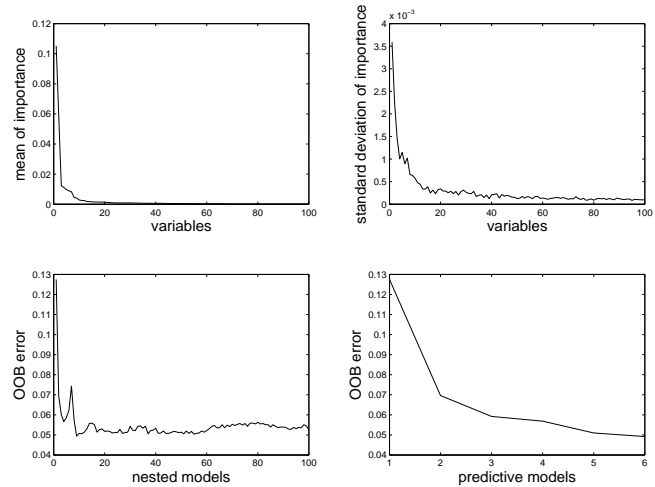}
         \end{center}
         \caption{Variable selection procedures for interpretation and prediction for Prostate data}
         \label{realvarselect}
\end{figure}

We apply the variable selection procedure on Prostate data. The
graphs of Figure \ref{realvarselect} are obtained as those of Figure
\ref{varselect}, except that for the RF procedure, we use
$ntree=2000$, $mtry=p/3$ and for the bottom left graph, we only plot
the $100$ most important variables for visibility. The procedure
leads to the same picture as previously, except for the OOB rate
along the nested models which is less regular. The key point is that
it selects $9$ variables for interpretation, and $6$ variables for
prediction. The number of selected variables is then very much
smaller than $p=6033$.

In addition, to examine the variability of the interpretation and
prediction sets  the global procedure is repeated five times on the
entire Prostate dataset. The five prediction sets are very close to
each other. The  number of prediction variables fluctuates between
$6$ and $10$, and $5$ variables appear in all sets. Among the five
interpretation sets, $2$ are identical and made of $9$ variables and
 the $3$ other are made of $25$ variables. The $9$ variables of the smallest
sets are present in all sets and the biggest sets (of size $25$)
have $23$ variables in common.

So, although the sets of variables are not identical for each run of the procedure,
they are not completely different. And in addition the most important variables are
included in all sets of variables.

\subsubsection{High dimensional classification}

We apply the global variable selection procedure on high dimensional
real datasets studied in Section \ref{classhigh}, and we want to get
an estimation of prediction error rates. Since these datasets are of
small size, we use a 5-fold cross-validation to estimate the error
rate. So we split the sample in $5$ stratified parts, each part is
successively used as a test set, and the remaining of the data is
used as a learning set. Note that the set of variables selected vary
from one fold to another. So, we give in Table
\ref{tabrealvarselect} the misclassification error rate, given by
the 5-fold cross-validation, for interpretation and prediction sets
of variables respectively. The number into brackets is the average
number of selected variables. In addition, one can find the original
error which stands for the misclassification rate given by the
5-fold cross-validation achieved with random forests using all
variables. This error is calculated using the same partition in 5
parts and again we use $ntree=2000$ and $mtry=p/3$ for all datasets.

\begin{table}[!ht]
         \begin{center}
     \begin{tabular}{c|c|c|c}
Dataset    & interpretation & prediction & original \\
\hline
Colon      & 0.16 (35)      & 0.20 (8)   & 0.14 \\
Leukemia   & 0 (1)          & 0 (1)      & 0.02 \\
Lymphoma   & 0.08 (77)     & 0.09 (12)   & 0.10 \\
Prostate   & 0.085 (33)     & 0.075 (8)  & 0.07 \\
     \end{tabular}
         \end{center}
         \caption{Variable selection procedure for four high dimensional
         real datasets. CV-error rate and into brackets the average number of selected variables}
         \label{tabrealvarselect}
\end{table}

The number of interpretation variables is hugely smaller than $p$,
at most tens to be compared to thousands. The number of prediction
variables is very small (always smaller than $12$) and the reduction can
be very important \emph{w.r.t} the interpretation set size. The
errors for the two variable selection procedures are of the same
order of magnitude as the original error (but a little bit larger).

We compare these results with the results obtained by Ben Ishak and
Ghattas (2008) (see tables 9 and 11 in \cite{BenIshak08}) which have
compared their method with 5 competitors (mentioned in the
introduction) for classification problems on these four datasets.
Error rates are comparable. With the prediction procedure, as
already noted in the introductory remark, we always select fewer
variables than their procedures (except for their method GLMpath
which select less than 3 variables for all datasets).

\subsection{Regression}

\subsubsection{A simulated dataset}

\begin{figure}[!ht]
         \begin{center}
         \includegraphics[width=0.7\textwidth]{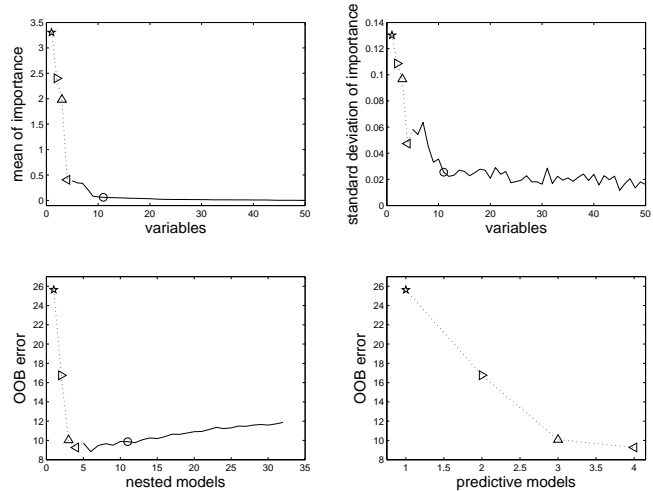}
         \end{center}
         \caption{Variable selection procedures for interpretation and prediction for Friedman1 data}
         \label{varselectfried1}
\end{figure}

We now apply the procedure to a simulated regression problem. We
construct starting from the Friedman1 model and adding noisy
variables as in Section \ref{reghigh}, a learning set of size
$n=100$ with $p=200$ variables. Figure \ref{varselectfried1}
displays the results of the procedure. The true variables of the
model (1 to 5) are respectively represented by
($\rhd,\bigtriangleup,\circ,\star,\lhd$).

The graphs are of the same kind as in classification problems. Note
that variable 3 is confused with noise and is not selected by the
procedure. This is explained by the fact that it is hardly
correlated with the response variable. The interpretation procedure
select the true variables except variable 3 and two noisy variables,
and the prediction set of variables contains only the true variables
(except variable 3). Again the whole procedure is stable in the
sense that several runs give the same set of selected variables.

In addition, we simulate a test set of the same size than the
learning set to estimate the prediction error. The test mean squared
error with all variables is about $19.2$, the one with the $6$
variables selected for interpretation is $12.6$ and the one with the
$4$ variables selected for prediction is $9.8$.

\subsubsection{Ozone data}

\begin{figure}[!ht]
         \begin{center}
       \includegraphics[width=0.5\textwidth]{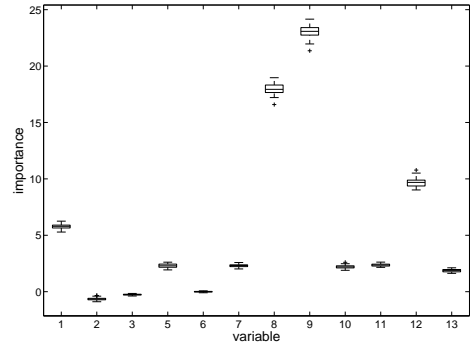}
         \end{center}
         \caption{Variable importance for Ozone data}
         \label{viozone}
\end{figure}

Before ending the paper, let us apply the entire procedure to the
ozone dataset. It consists of $n=366$ observations of the daily
maximum one-hour-average ozone together with $p=12$ meteorologic
explanatory variables. Let us first examine, in Figure \ref{viozone}
the VI obtained with RF procedure using $mtry=p/3=4$ and
$ntree=2000$.

From the left to the right, the 12 explanatory variables are
1-Month, 2-Day of month, 3-Day of week, 5-Pressure height, 6-Wind
speed, 7-Humidity, 8-Temperature (Sandburg), 9-Temperature (El
Monte), 10-Inversion base height, 11-Pressure gradient, 12-Inversion
base temperature, 13-Visibility.

Three very sensible groups of variables appear from the most to the
least important. First, the two temperatures (8 and 9), the
inversion base temperature (12) known to be the best ozone
predictors, and the month (1), which is an important predictor since
ozone concentration exhibits an heavy seasonal component. A second
group of clearly less important meteorological variables: pressure
height (5), humidity (7),
  inversion base height (10), pressure gradient (11) and visibility (13).
  Finally three unimportant variables: day of month (2), day of week (3)
  of course and more surprisingly wind speed (6). This last fact is
  classical: wind enter in the model only when ozone pollution
  arises, otherwise wind and pollution are uncorrelated (see for
  example Cheze et al. (2003) \cite{Cheze03} highlighting this phenomenon using
  partial estimators).

Let us now examine the results of the selection procedures.

\begin{figure}[!ht]
         \begin{center}
         \includegraphics[width=0.7\textwidth]{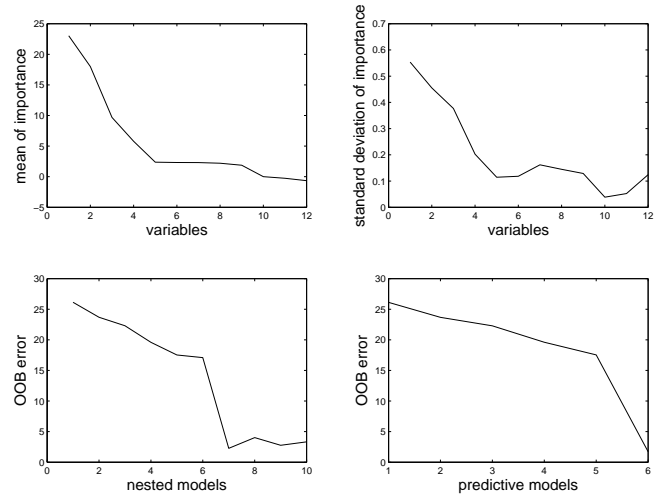}
         \end{center}
         \caption{Variable selection procedures for interpretation and prediction for Ozone data}
         \label{varselectozone}
\end{figure}

After the first elimination step, the 2 variables of negative
importance are canceled, as expected.

Therefore we keep 10 variables for interpretation step and then the
model with 7 variables is then selected and it contains all the most
important variables: (9 8 12 1 11 7 5).

For the prediction procedure, the model is the same except one more
variable is eliminated: humidity (7) .

In addition, when different values for $mtry$ are considered, the
most important 4 variables (9 8 12 1) highlighted by the VI index,
are selected and appear in the same order. The variable 5 also
always appears but another one can appear after of before.

\section{Discussion}

Of course, one of the main open issue about random forests is to
elucidate from a mathematical point of view its exceptionally
attractive performance. In fact, only a small number of references
deal with this very difficult challenge and, in addition to bagging
theoretical examination by B\"{u}lmann and Yu (2002)
\cite{Buhlmann02}, only purely random trees, a simple version of
random forests, is considered. Purely random trees have been
introduced by Cutler and Zhao (2001) \cite{Cut01} for classification
problems and then studied by Breiman (2004) \cite{Breiman04}, but
the results are somewhat preliminary. More recently Biau \emph{et
al.} (2008) \cite{Biau08} obtained the first well stated consistency
type results.

From a practical perspective, surprisingly, this simplified and
essentially not data-driven strategy seems to perform well, at least
for prediction purpose (see Cutler and Zhao 2001 \cite{Cut01}) and,
of course, can be handled theoretically in a easier way.
Nevertheless, it should be interesting to check that the same
conclusions hold for variable importance and variable selection
tasks.

In addition, it could be interesting to examine some variants of
random forests which, at the contrary, try to take into account more
information. Let us give for example two ideas. The first is about
pruning: why pruning is not used for individual trees? Of course,
  from the computational point
of view the answer is obvious and for prediction performance,
averaging eliminate the negative effects of individual overfitting.
But from the two other previously mentioned statistical problems, prediction
and variable selection, it remains unclear.
The second remark is about the random feature
selection step. The most widely used version of RF selects randomly
$mtry$ input variables according to the discrete uniform
distribution. Two variants can be suggested: the first is to select
random inputs according to a distribution coming from a preliminary
ranking given by a pilot estimator; the second one is to adaptively
update this distribution taking profit of the ranking based on the
current forest which is then more and more accurate.

These different future directions, both theoretical and practical,
will be addressed in the next step of the work.

%

\section{Appendix}

In the sequel, information about datasets retrieved from the R
package \texttt{mlbench} can be found in the corresponding
description file. \\

\textbf{Standard problems, $n>>p$:}

\begin{itemize}
  \item Binary classification
  \begin{itemize}
    \item Real data sets\footnote[3]{from the R package \texttt{mlbench}}
  \begin{itemize}
    \item Ionosphere $(n=351,p=34)$
    \item Diabetes, \texttt{PimaIndiansDiabetes2} $(n=768,p=8)$
    \item Sonar $(n=208,p=60)$
    \item Votes, \texttt{HouseVotes84} $(n=435,p=16)$
  \end{itemize}
      \item Simulated data sets\footnotemark[3]
    \begin{itemize}
      \item Ringnorm, \texttt{mlbench.ringnorm} $(n=200,p=20)$
      \item Threenorm, \texttt{mlbench.threenorm} $(n=200,p=20)$
      \item Twonorm, \texttt{mlbench.twonorm} $(n=200,p=20)$
    \end{itemize}

  \end{itemize}

  \item Multiclass classification
  \begin{itemize}
    \item Real data sets\footnotemark[3]
   \begin{itemize}
     \item Glass $(n=214,p=9,c=6)$
     \item Letters, \texttt{LetterRecognition} $(n=20000,p=16,c=26)$
     \item Sat-images, \texttt{Satellite} $(n=6435,p=36,c=6)$
     \item Vehicle $(n=846,p=18,c=4)$
     \item Vowel $(n=990,p=10,c=11)$
   \end{itemize}

    \item Simulated data sets\footnotemark[3]
   \begin{itemize}
     \item Waveform, \texttt{mlbench.waveform} $(n=200,p=21,c=3)$
   \end{itemize}

  \end{itemize}

  \item Regression
  \begin{itemize}
    \item Real data sets\footnotemark[3]
    \begin{itemize}
     \item BostonHousing $(n=506,p=13)$
     \item Ozone $(n=366,p=12)$
     \item Servo $(n=167,p=4)$
    \end{itemize}

    \item Simulated data sets\footnotemark[3]
    \begin{itemize}
     \item Friedman1, \texttt{mlbench.friedman1} $(n=300,p=10)$
     \item Friedman2, \texttt{mlbench.friedman2} $(n=300,p=4)$
     \item Friedman3, \texttt{mlbench.friedman3} $(n=300,p=4)$
    \end{itemize}

  \end{itemize}
\end{itemize}

\textbf{High dimensional problems, $n<<p$:}

\begin{itemize}
  \item Binary classification
  \begin{itemize}
    \item Real data sets\footnote[4]{see http://ligarto.org/rdiaz/Papers/rfVS/randomForestVarSel.html}
  \begin{itemize}
    \item Adenocarcinoma $(n=76,p=9868)$, see Ramaswamy \emph{\emph{et al.}} (2003) \cite{Ramaswamy03}
    \item Colon $(n=62,p=2000)$, see Alon \emph{\emph{et al.}} (1999) \cite{Alon99}
    \item Leukemia $(n=38,p=3051)$: see Golub \emph{\emph{et al.}} (1999) \cite{Golub99}
    \item Prostate $(n=102,p=6033)$: see Singh \emph{\emph{et al.}} (2002) \cite{Singh02}

  \end{itemize}
      \item Simulated data sets\footnote[5]{see description in section \ref{classhigh}}
    \begin{itemize}
      \item toys data $(n=100,100\leq p\leq 1000)$, see Weston \emph{\emph{et al.}} (2003) \cite{Weston03}
    \end{itemize}

  \end{itemize}

  \item Multiclass classification
  \begin{itemize}
    \item Real data sets\footnotemark[4]
    \begin{itemize}
     \item Brain $(n=42,p=5597,c=5)$, see Pomeroy \emph{\emph{et al.}} (2002) \cite{Pomeroy02}
     \item Breast, \texttt{breast.3.class} $(n=96,p=4869,c=3)$, see van't Veer \emph{\emph{et al.}} (2002) \cite{Veer02}
     \item Lymphoma $(n=62,p=4026,c=3)$, see Alizadeh (2000) \cite{Alizadeh00}
     \item Nci $(n=61,p=6033,c=8)$, see Ross \emph{\emph{et al.}} (2000) \cite{Ross00}
     \item Srbct $(n=63,p=2308,c=4)$, see Khan \emph{\emph{et al.}} (2001) \cite{Khan01}
    \end{itemize}

  \end{itemize}

  \item Regression
  \begin{itemize}
    \item Real data sets\footnote[6]{from the R package \texttt{chemometrics}}
    \begin{itemize}
     \item PAC $(n=209,p=467)$
    \end{itemize}

    \item Simulated data sets\footnotemark[3]
    \begin{itemize}
     \item Friedman1, \texttt{mlbench.friedman1} $(n=100,100\leq p\leq 1000)$
    \end{itemize}

  \end{itemize}
\end{itemize}


\end{document}